\documentclass[sigconf,balance=true]{acmart}

\setcopyright{rightsretained} \acmYear{2021} \acmISBN{} \acmDOI{} \acmConference[PatentSemTech]{2nd Workshop on Patent Text Mining and Semantic Technologies (PatentSemTech2021) co-located with the 44th International ACM SIGIR Conference on Research and Development in Information Retrieval}{July 15th, 2021}{online} \makeatletter \renewcommand{\@copyrightpermission}{}
\renewcommand{\@copyrightowner}{for this paper by its authors. Use permitted under Creative Commons License Attribution 4.0 International (CC BY 4.0). CEUR Workshop Proceedings (CEUR-WS.org)} \makeatother 

\AtBeginDocument{%
  \providecommand\BibTeX{{%
    \normalfont B\kern-0.5em{\scshape i\kern-0.25em b}\kern-0.8em\TeX}}}

\usepackage{svg}
\usepackage{xcolor,listings}

\begin{document}

\title{Linguistically Informed Masking for Representation Learning in the Patent Domain}

\author{Sophia Althammer}
\affiliation{%
  \institution{TU Vienna}
  \city{Vienna}
  \country{Austria}
}
\email{sophia.althammer@tuwien.ac.at}

\author{Mark Buckley}
\affiliation{
  \institution{Siemens AG}
  \city{Munich}
  \country{Germany}}
\email{mark.buckley@siemens.com}

\author{Sebastian Hofst{\"a}tter}
\affiliation{
  \institution{TU Vienna}
  \city{Vienna}
  \country{Austria}}
\email{sebastian.hofstaetter@tuwien.ac.at}


\author{Allan Hanbury}
\affiliation{
  \institution{TU Vienna}
  \city{Vienna}
  \country{Austria}
}
\email{allan.hanbury@tuwien.ac.at}

\renewcommand{\shortauthors}{Althammer, et al.}

\begin{abstract}
  Domain-specific contextualized language models have demonstrated substantial effectiveness gains for domain-specific downstream tasks, like similarity matching, entity recognition or information retrieval. However successfully applying such models in highly specific language domains requires domain adaptation of the pre-trained models. In this paper we propose the empirically motivated \emph{Linguistically Informed Masking} (LIM) method to focus domain-adaptative pre-training on the linguistic patterns of patents, which use a highly technical sublanguage. We quantify the relevant differences between patent, scientific and general-purpose language and demonstrate for two different language models (BERT and SciBERT) that domain adaptation with LIM leads to systematically improved representations by evaluating the performance of the domain-adapted representations of patent language on two independent downstream tasks, the IPC classification and similarity matching. We demonstrate the impact of balancing the learning from different information sources during domain adaptation for the patent domain. We make the source code as well as the domain-adaptive pre-trained patent language models publicly available at \newline https://github.com/sophiaalthammer/patent-lim.
\end{abstract}



\maketitle

\section{Introduction}
Large scale language models, pre-trained on corpora of general purpose language \cite{devlin2018bert}, provide effective representations for text documents, which improve the performance on a variety of downstream tasks including information retrieval, information extraction and similarity matching \cite{beltagy2019scibert,nogueira2019passage,Krestel2021}. The representations of contextualized language models are used in production systems in the web and news domains to include semantic knowledge for solving tasks based on text input like search or automated classification \cite{linkgoogle,Aristodemou2018patentsurvey}.

In this paper we propose a novel domain-adaption \emph{linguistically informed masking} pre-training method for BERT-style language models. We show its suitability for the patent domain and demonstrate its effectiveness for representation learning on patent language.
We publish our BERT-based language model pre-trained on patent language data to make our work readily available for the community.

The suitability of the representations with respect to a given downstream task relies on the assumption that the language of the documents of that downstream task comes from the same or similar distribution as data of the language model. Otherwise domain-adaptive pre-training becomes necessary, as the further the language of the downstream task is from the distribution of the pre-training language, the less relevant information is encoded in the representations \cite{gururangan-etal-2020-dontstoppretraining,han2019unsupervised}. Therefore we investigate in this paper: 

\begin{itemize}
    \item[\textbf{RQ1}] Which BERT-like pre-trained language model is best suited to representing patent language?
\end{itemize}

We compare the language models BERT \cite{devlin2018bert} (pre-trained on general purpose language) and SciBERT \cite{beltagy2019scibert} (pre-trained on scientific language for representing patent language). We perform domain-adaptive pretraining with both models and evaluate the resulting representations on two independent patent-related downstream tasks: IPC classification and similarity matching. Here we find that the downstream task performance of the SciBERT based representations outperforms the BERT-based representations for both patent-related tasks. Furthermore we reason that this is due to the more fine-grained tokenization of the patent language by the Sci- BERT model than by the BERT models' tokenization.\newline
Patent language contains linguistic patterns which differ from general purpose or scientific language \cite{oostdijk2010} and is characterised by the frequent use of technical terms and novel multi-word expressions, as well as long sentences, chained conjunctions and large noun phrases, as shown in Figure~\ref{fig:example_claim}. In order to take these linguistic characteristics into account, we propose the domain-adaptive pre-training scheme for BERT-like language models: \textbf{linguistically informed masking}. Linguistically informed masking shifts the masking probabilities in domain-adaptive pre-training towards the highly informative multi-word terms in patent language. As multi-word terms are contained in the noun chunks, the degree of shifting the masking towards the multi-word terms is here controlled with a noun chunk masking probability $p_{nc}$. We investigate:

\begin{itemize}
    \item[\textbf{RQ2}] How does domain-adative pre-training with linguistically informed masking influence the representations of patent language?
\end{itemize}

For BERT and SciBERT we compare domain-adaptive pre-training with and without linguistically informed masking with shifting the masking to $100\%$ and to $75\%$ to the noun chunks. In order to evaluate the effectiveness of the different representations for patent language, we evaluate and compare them for the two independent downstream tasks of IPC classification and similarity matching. Here we find that linguistically informed masking does improve the overall downstream performance, thus we reason that shifting the masking probabilities towards the noun chunks improves the representations of patent language. Furthermore we investigate the degree of shifting the masking probability $p_{nc}$ towards the noun chunks.

\begin{figure}
    What is claimed is: 
    1. A \textbf{hydraulic regeneration deactivation valve} to react to a pressure and to deactivation regeneration of a hydraulic cylinder,... 
    \vspace{0.1cm}
    \vspace{-0.2cm}
    \caption{An example of a patent claim}
    \label{fig:example_claim}
\end{figure}

Our contributions are as follows:

\begin{itemize}
    \item We investigate domain adaptive pre-training of a general purpose language model (BERT) and of a scientific language model (SciBERT) for representing patent language and find that SciBERT is more suitable to represent patent language. We investigate two factors for that: the language pre-training corpus as well as the tokenization
    \item We propose linguistically informed masking for domain-adaptive pre-training for patent language and demonstrate effectiveness gains with representations of patent language learned with linguistically informed masking for IPC classification and similarity matching
    \item We make the source code as well as the domain-adapted pre-trained language models available at \newline https://github.com/sophiaalthammer/patent-lim
\end{itemize}

\section{Related work}

\paragraph{Representation learning}
Learning general word representations continues to be an active research area, from word-level representations \cite{mikolov2013distributed, glove} up to  pretrained language models \cite{elmo, openaigpt2, ulmfit, devlin-etal-2019-bert}.
In particular the BERT language model \cite{devlin-etal-2019-bert} delivers generally applicable, syntactically and semantically informative embeddings which have advanced the state-of-the-art performance on a variety of different downstream tasks. The extensive and varied further research results based on the BERT model \cite{m2019spanbert, liu2019roberta, sun2019ernie, lan2019albert} show the generality and flexibility of the representations.
\citet{Sun2019ERNIE1, sun2019ernie} use entity-level and phrase-level masking to achieve state-of-the-art results on Chinese. \citet{m2019spanbert} explore the effects of different static masking schemes for the BERT pre-training and find that random span masking is the best for learning general-purpose language.

\paragraph{Domain adaptation of language models}
There is a suite of BERT-like domain-specific models which have been fine tuned for, e.g., the social media
\cite{han2019unsupervised}, biomedical \cite{biobert}, clinical \cite{clinicalbert}, legal \cite{chalkidis-etal-2020-legalbert} or scientific domains \cite{beltagy2019scibert}. They show that the domain adaptive fine-tuning on the same language modelling tasks already leads to more informative representations of the respective domain and therefore to better performance on downstream tasks. \citet{beltagy2019scibert} demonstrate by training a BERT language model from scratch on scientific language and with a trained, scientific vocabulary that the suitability of the tokenization to the domain language is an important parameter for good representations of that language. However \citet{gururangan-etal-2020-dontstoppretraining} demonstrate that domain-adaptive pre-training is crucial for specific domains and for the performance on downstream tasks. \citet{sebastianpatentretrieval} demonstrate the use of retrofitting \cite{faruqui-etal-2015-retrofitting} for Word2Vec  \cite{mikolov2013distributed} patent embeddings for patent retrieval.

\paragraph{Natural language processing in the patent domain}
The use of machine learning and deep learning methods for patent analysis is a vibrant research area \cite{Krestel2021,Aristodemou2018patentsurvey} with application in technology forecasting, patent retrieval \cite{althammer2021crossdomain,clefipsummary13}, patent text generation \cite{lee2020patenttransformer2} or litigation analysis.
There has been much research on the patent domain language which shows that the sections in patents constitute different genres depending on their legal or technical purpose \cite{risch2019}. Furthermore the vocabulary of patent language is highly specific \cite{lupu, oostdijk2010} and contains special multi-word terms which are 
novel constructions from commonly used words and which are characteristic of patent language, as in Figure~\ref{fig:example_claim} \cite{verberne2010, dhondt2009}.
The citations of patents are a frequent subject of research and they are used to explore the similarity of cited patents \cite{citationspatent, citationlinkprediction, risch2020patentmatch}. The classification of patents with the IPC tags, which determine a hierarchical topic category of the patent, is a well known downstream task \cite{deeppatent, lee2019patentbert}.
The manual curation of patent metadata by the patent offices provides abundant labelled data for NLP research, however the tasks of IPC classification and similarity matching can not yet be considered solved, and thus are sufficiently difficult tasks for comparing the capabilities of representations for patent language.

\section{Linguistically informed masking}

Here we motivate the domain adaptive pre-training method of linguistically informed masking for learning patent representations and we define and introduce linguistically informed masking for BERT-like language models.

\subsection{Linguistic motivation}
\label{chap:linguisticanalysis}

Our goal is to learn better language representations for the patent domain. As the language models which are the subject of our research are pre-trained on a different language domain, we aim to quantify the difference between the pre-training language and the target language, here patent language, in terms of their linguistic patterns. Therefore we examine the general-purpose and scientific language on which BERT and SciBERT are pre-trained, respectively, as these models will serve as pre-trained language models which we adapt to patent language.

One of the main distinct characteristics of patent language is the use of constructed multi-word terms \cite{oostdijk2010} such as ``a disk-shaped suspension-type insulator'' or ``a non-transitory computer-readable medium''.
These multi-word terms are contained in noun chunks, therefore we analyze the length and appearance of noun chunks in patent language compared to general-purpose and scientific language. As representatives from each language domain we choose a sample of $600$ articles from the Wikitext raw training dataset \cite{merity2016pointer} and $1,000$ abstracts from the Semantic Scholar research corpus \cite{Ammar2018ConstructionOT}, which correspond to the pre-training data of BERT and SciBERT.

We identify the noun chunks using the Spacy natural language toolkit\footnote{https://github.com/explosion/spaCy/} after removing tabs and multiple whitespaces as well as mathematical formulas from the raw text. We remove noun chunks which are longer than $10$ words as we see that these cases are enumerations when analyzing the noun chunks.

\begin{figure}
    \centering
    \includegraphics[width=0.50\textwidth]{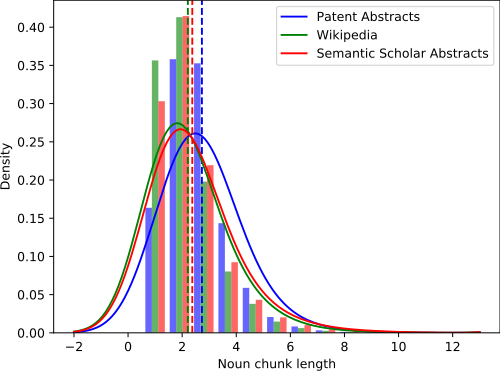}
    \caption{Distribution of noun chunk length in patent and Semantic Scholar abstracts and Wikipedia: patent language contains on average longer noun chunks with domain specific information signals}
    \label{fig:nounphrasecomparison}
\end{figure}

Figure \ref{fig:nounphrasecomparison} shows the distribution of noun chunk length of the different language domains, with the average length in dashed vertical lines.
We observe a significant difference of the distribution of the noun chunk length in the patent language (mean: $2.73$; sd: $1.27$) compared to general-purpose ($2.21$;$1.25$) or scientific language ($2.37$;$1.36$) as there are more long noun chunks in patent language (K-S test: $p<.001$ for all three language combinations). Therefore we conclude that patent language contains longer noun chunks than general-purpose or scientific language and the longer noun chunks in patent language are constructed using novel combinations of common nouns. Consequently the BERT and SciBERT models are not trained to 
generate optimal representations of these noun chunks.

Considering that the noun chunks contain domain-specific information signals in the form of multi-word-terms and technical terms \cite{oostdijk2010}, this motivates us 
to focus on learning the linguistic peculiarities of the patent domain contained in the noun chunks explicitly during domain-adaptative fine-tuning. Hence we propose linguistically informed masking for domain-adaptive fine-tuning of BERT-like language models.

\begin{table}[t]
\small
\centering
    \begin{tabular}{@{}lcc@{}}
    \toprule
     & Wikipedia & USPTO13M\\
    \midrule
    $p(y_{jk}=1)$ & 0.499 & 0.507\\
    $p(y_{jk}=0)$ & 0.501 & 0.493\\
    \bottomrule
    \end{tabular}
    \caption{Probability that a given token $k$ in sequence $j$ is in a noun chunk for two datasets}
    \label{table:noun_chunk_dist}
\end{table}

\subsection{Masked Language Modelling (MLM)}

The BERT language model \cite{devlin-etal-2019-bert} is designed to learn bidirectional representations for language and is jointly pre-trained on the tasks of masked language modelling (MLM) and next sentence prediction (NSP), with two different additional layers based on the output of its transformer network \cite{vaswani2017attention}. 
In the next sentence prediction task the model is given two sentences of a text and it has to predict if the second sentence is the next sentence in the original text.
In masked language modelling $15\%$ of the tokens in each sequence are masked out and the model predicts the true token, which is inspired by the Cloze task \cite{Taylor1953ClozePA}.
Solving these pre-training tasks requires the representations to capture syntactic and semantic characteristics of the language and therefore this task enables the language model to learn linguistic representations.\newline
In this section we describe the novel domain-adaptive pre-training method of \emph{linguistically informed masking} (LIM) for BERT-like language models, which adapts the MLM task in order to focus the model towards learning specific linguistic information of the domain.\newline
We first give a formal definition of the MLM task.
Let $B$ be the number of training sequences consisting of encoded sentences in one training batch and let $max\textunderscore pred$ be the number of masked positions where the original token needs to be predicted. Then the loss $L$ of the MLM task is defined as cross-entropy between the predictions $p\textunderscore mlm_{ij}$ and the label $l\textunderscore mlm_{ij}$ as
\begin{align*}
    L := \frac{\sum\limits_{j=1}^{B} \sum\limits_{i=1}^{max\textunderscore pred} - (\log(p\textunderscore mlm_{ij})^{T} \ l\textunderscore mlm_{ij}) w_{ij}} {\sum\limits_{j=1}^{B} \sum\limits_{i=1}^{max\textunderscore pred} w_{ij}}
\end{align*}
for each position $i$ and for each sequence $j$.
$w_{ij}$ is a weight for a padding mechanism 
 in case fewer than $max\textunderscore pred$ positions are masked in the sequence. The predictions $p\textunderscore mlm_{ij} \in \mathbb{R}^V$ are a probability distribution over the whole vocabulary with size $V$ and the label is a one-hot encoding of the masked token.
The predictions $p\textunderscore mlm_{ij}$ are the output of the MLM layer and are defined as
\begin{align*}
    p\textunderscore mlm_{ij} := \ softmax(W_{mlm} \ X_{ij} \ + \ b_{mlm})
\end{align*}
where $W_{mlm} \in \mathbb{R}^{V \times H}$ and $b_{mlm} \in \mathbb{R}^{V}$ are the weights and biases of the MLM output layer and $X_{ij} \in \mathbb{R}^{H}$ is the final hidden vector of the masked position $i$ with dimensionality $H$.
For a sequence $j$ we get the final hidden vectors of the masked positions $X_{j} \in \mathbb{R}^{H \times max\textunderscore pred}$ with
\begin{align*}
    X_j = \ T_j M_j \ \ \forall j=1,.., B.
\end{align*}
Here $T_j \in \mathbb{R}^{H \times S}$ are the final outputs of sequence $j$ of the BERT model with the input sequence length $S$ and the masking matrix $M_j \in \{0,1\}^{S \times max\textunderscore pred}$, which shows the masked positions.
The masking matrix $M^j$ for a sequence $j$ consists of one-hot vectors for each masked position $n$ with $n=1,.., max\textunderscore pred$:
\begin{align*}
    M^j_{kn} := \begin{cases}
      1, & \text{if token $k$ is masked in $n$th position}\\
      0, & \text{otherwise.}
\end{cases}
\end{align*}

\subsection{Linguistically informed masking pre-training method (LIM)}
\label{chap:limanalysis}

With the linguistically informed masking method we aim to give the model the flexibility to focus on learning specific linguistic characteristics of the language, namely the noun chunks in the patent domain. Therefore we propose the linguistically informed masking method where $p_{nc} \in [0,1]$ of the masked tokens belong to a noun chunk and $1-p_{nc}$ to a non-noun chunk.
We realize this by modifying the masking matrix $M^j \in \{0, 1\}^{S \times max\textunderscore pred}$ of sequence $j$ depending on 
\begin{align*}
    y_{jk} := \begin{cases}
      1, & \text{if token $k$ belongs to a noun chunk} \\
      0, & \text{otherwise}
\end{cases}
\end{align*}
which indicates whether token $k$ of sequence $j$ belongs to a noun chunk or not.
The LIM masking matrix $\hat{M}^j \in \{0, 1\}^{S \times max\textunderscore pred}$ is constructed so that it contains with a probability of $p_{nc}$ only masked tokens $k$ for which $y_{jk}=1$, and with a probability of $1-p_{nc}$ only masked tokens for which $y_{jk}=0$.
With this construction, in $p_{nc}$ of the sequences only tokens that belong to a noun chunk are masked, and in $1-p_{nc}$ of the sequences only tokens of non-noun chunks are masked.\newline
In order to examine the differences of the pre-training methods of MLM and LIM we analyze the overall noun chunk distribution for tokens in the Wikitext raw dataset \cite{merity2016pointer}, which corresponds to the pre-training domain of the BERT model, and the USPTO13M patent dataset, which is shown in Table~\ref{table:noun_chunk_dist}.
Because masking in MLM is random and around half of the tokens in both datasets are part of noun chunks, around half of the masked positions will belong to a noun chunk and half will not.
This means that with MLM fine-tuning on patent documents, the influence of tokens in noun chunks would be approximately equally weighted, despite the importance of noun chunks in patent language.
With LIM however we can control the influence of noun chunks via the parameter $p_{nc}$.
To increase this influence the noun chunk masking probability must be $p_{nc} > p(y_{jk} = 1)$, which means that we choose $p_{nc} > 0.507$ for the patent domain. With $p_{nc}=0.507$ LIM reduces to MLM as a special case, as the probability of masking out a noun chunk in LIM is then the same as in MLM.

We examine the impact of LIM compared to MLM with the choice of $p_{nc} = 0.75$. The probability of masking the token $k$, which is in a noun chunk in sequence $j$, with the MLM task in the patent domain is

\begin{align*}
    p(M^j_{kn} = 1&| y_{jk} = 1) = \frac{p ( M^j_{kn} = 1 \wedge y_{jk} = 1)}{p(y_{jk} = 1)}\\
    =& \frac{p(M^j_{kn}=1) * p(y_{jk} = 1)}{p(y_{jk} = 1)} = 0.15
\end{align*}

as the masking is independent of the noun chunks.
For LIM the probability of masking a noun chunk token is

\begin{align*}
    p(&\hat{M}^j_{kn} = 1| y_{jk} = 1) = \frac{p ( \hat{M}^j_{kn} = 1 \wedge y_{jk} = 1)}{p(y_{jk} = 1)} \\
    & = \frac{p(\hat{M}^j_{kn}=1) * p_{nc}}{p(y_{jk} = 1)} = \frac{0.15 * 0.75 }{0.507} = 0.22
\end{align*}

This shows how we influence the probability of masking a token $k$, which is in a noun chunk, with the parameter $p_{nc}$. 
With $p_{nc}=0.507$ LIM reduces to MLM as a special case. In summary, the LIM parameter $p_{nc}$ controls the probability of masking noun chunk tokens.

\section{Experiment design}
\label{chap:experiments}

Our experiments investigate which BERT-like pre-trained language model is best suited to represent patent language (RQ1) as well as the influence of domain-adaptive pre-training with linguistically informed masking for the representations of patent language (RQ2).\newline
To evaluate these questions we do domain-adaptive pre-training on patent language with BERT and SciBERT with either the MLM or LIM pre-training method. We compare the vanilla model without domain-adaptive pre-training, the MLM and LIM domain-adaptive pre-training for BERT and SciBERT for representing patent language.
To assess the quality of the resulting representations we evaluate the performance of the representations for the two independent, patent-related downstream tasks of IPC classification and similarity matching.\footnote{We show the tasks' independence in Appendix~\ref{appendix:independeceipccit}.}

\subsection{Data}

We leverage the patent corpus from the Google Patents Public Datasets \footnote{https://console.cloud.google.com/bigquery?p=patents-public-data} on BigQuery with the query in Appendix \ref{appendix:a}. The corpus, which we will refer to as USPTO13M, consists of $13$ million granted utility patents in English with title, abstract, claims and description. The title contains on average $8$ words, the abstract $112$ words, the claims $1067$ and the descriptions $9539$ words. We also retrieve metadata like the filing date and the IPC tags, which are a consistent, hierarchical topic categorization of the patents and which are assigned by patent examiners \cite{wipo_ipc}. Our corpus contains $738$ different IPC tags on the subclass level of the tags. The patents also contain citations referring to other previously published patents which the current patent is related to.
Of primary importance are category ``X'' citations, which encode close technical relatedness. Category ``X'' is applicable ``where a document is such that when taken alone, a claimed invention cannot be considered novel'' \cite{citationtypes}. We will use the citations of a patents as similarity indication of two patents.

\subsection{Domain adaptive pre-training on patent language}

Following the definition of \citet{deftransferlearning} for transfer learning, we define the domain adaptive pre-training from the source domain Wikipedia or the source domain of scientific language with the source tasks of MLM and NSP to the target domain of patent language with the target task of MLM or LIM and NSP.
Here we take the unsupervised domain-adaptive pre-training dataset as the title, abstract, claims and descriptions of $320k$ patent documents from the USPTO13M corpus containing $3.3$ billion words, similar to the size of the BERT and SciBERT pre-training datasets.
We remove tabs, multiple whitespaces and mathematical formulas. 
We choose the $BERT_{BASE}$ cased model with $110M$ parameters as initialisation to match the size of the SciBERT model.
As the learning rate is a sensitive parameter that is data- and task-specific, we perform a grid search over $1*10^{-4}$, $5*10^{-5}$, $2*10^{-5}$, $1*10^{-5}$ as learning rate candidates following \citet{beltagy2019scibert} and \citet{clinicalbert} and choose the optimal learning rate for each configuration.
We choose the same hyperparameters as were used in the pre-training phase for each model. The hyperparameters, including learning rates, can be found in Appendix \ref{appendix:domainfinetunelearningrate}.
We carry out domain adaptive pre-training of the BERT and SciBERT model for $100k$ steps following \citet{biobert}, which equates to one epoch of the domain adaptive pre-training dataset.\newline
For LIM domain adaptive pre-training we consider two values for our domain-dependent hyperparameter $p_{nc}\in \{0.75, 1.0\}$ as our analysis in section \ref{chap:limanalysis} has shown that $p_{nc}=0.507$ reduces to MLM pre-training and as we want to investigate the optimal weighting of LIM hyperparameter $p_{nc}$. 

\subsection{Fine-tuning on patent-related downstream tasks}

Our goal is to compare the performances of the different representations on the downstream tasks rather than to maximise the absolute performances. As two independent patent-related downstream tasks for evaluating the quality of the representations of patent language we choose IPC classification and similarity matching of two given patents.
As baselines we choose the BERT and SciBERT vanilla models which are not domain adapted to the patent domain, as well as a convolutional neural network for sentence classification \cite{kim-2014-convolutional} based on word2vec representations \cite{mikolov2013distributed}.

Following the definition of \citet{deftransferlearning} the vanilla BERT model is transferred from the source domain Wikipedia and the vanilla SciBERT model is transferred from the scientific source domain, to the patent domain with the target tasks of IPC classification or similarity matching. 
Therefore the transfer learning problem involves both a domain and a task shift.

For the BERT model which is fine-tuned on the patent domain with the MLM task the transfer is defined as only a task adaptation from the source task of MLM or LIM and NSP to the target task of IPC classification or similarity matching on the same domain of patents. 
The fine-tuning of the BERT LIM0.75 and the LIM1 configuration are defined analogously for the SciBERT-based models.
Overall we fine-tune and evaluate $8$ different pre-trained models on the IPC classification and the similarity matching tasks.

\begin{table*}[]
\small
\centering
\begin{tabular}{@{}lcccccccc@{}}
\toprule
        IPC classification  & \multicolumn{2}{c}{Accuracy} & \multicolumn{2}{c}{Precision} & \multicolumn{2}{c}{Recall} & \multicolumn{2}{c}{F1 Score} \\
\cmidrule(lr){2-3} \cmidrule(lr){4-5} \cmidrule(lr){6-7} \cmidrule(lr){8-9}
          & 160k & 320k & 160k & 320k & 160k & 320k &  160k &  320k \\
\midrule
Word2Vec + CNN    & 0.2600  & 0.2900 & 0.2600 & 0.2700 & 0.2600  & 0.2900  & 0.1900 & 0.2100 \\
\midrule
\textbf{BERT} & & & & & & & & \\
\midrule
VanillaBERT    & 0.5413  & 0.5779    & 0.5244   & 0.5663   & 0.5413  & 0.5779  & 0.5163  & 0.5605 \\
\midrule
   Domain adapted & & & & & & & & \\
\midrule
$p_{nc}=0.50$ (MLM) & $0.5503^\dagger$   & $0.5813^\dagger$  & $0.5275^\dagger$ & $\textbf{0.5744}^\dagger$  & $0.550^\dagger$ & $0.5813^\dagger$ & $0.5250^\dagger$  & $\textbf{0.5651}^\dagger$\\
$p_{nc}=0.75$ (LIM0.75, ours)  & $\textbf{0.5531}^\dagger$ & $\textbf{0.5820}^\dagger$ & $\textbf{0.5296}^\dagger$ & $0.5703^\dagger$ & $\textbf{0.5531}^\dagger$ & $\textbf{0.5820}^\dagger$ & $\textbf{0.5279}^\dagger$  & $0.5647^\dagger$  \\
$p_{nc}=1.00$ (LIM1, ours)   & $0.5472^\dagger$  & $0.5790^\dagger$ & $0.5227^\dagger$ & $0.5700^\dagger$  & $0.5472^\dagger$ & $0.5790^\dagger$ & $0.5218^\dagger$  & $0.5614^\dagger$ \\
\midrule
   \textbf{SciBERT} & & & & & & & & \\
\midrule
VanillaSciBERT & 0.5604& 0.5864& 0.5422& 0.5782& 0.5604& 0.5864& 0.5356& 0.5709\\
\midrule
   Domain adapted & & & & & & & & \\
\midrule
$p_{nc}=0.50$ (MLM) & $0.5636^\dagger$& $0.5909^\dagger$& $0.5414^\dagger$& $0.5800^\dagger$& $0.5636^\dagger$& $0.5909^\dagger$& $0.5386^\dagger$& $0.5738^\dagger$\\
$p_{nc}=0.75$ (LIM0.75, ours)  & $\textbf{0.5693}^\dagger$& $0.5927^\dagger$& $0.5486^\dagger$& $0.5821^\dagger$& $\textbf{0.5693}^\dagger$ & $0.5927^\dagger$ &$\textbf{0.5449}^\dagger$& $0.5760^\dagger$\\
$p_{nc}=1.00$ (LIM1, ours) &$0.5626^\dagger$& $\textbf{0.5955}^\dagger$& $\textbf{0.5493}^\dagger$& $\textbf{0.5840}^\dagger$& $0.5626^\dagger$&$\textbf{0.5955}^\dagger$& $0.5420^\dagger$ & $\textbf{0.5778}^\dagger$\\
\bottomrule
\end{tabular}
\caption{Accuracy, precision, recall and F1-score of IPC classification on the test set for BERT and SciBERT without and with domain-adaptive pre-training with MLM or LIM ($p_{nc}=0.75/1.00$), $^\dagger$ indicates statistically significant difference to Vanilla baseline, $\alpha=0.05$}
\label{table:evalipc}
\end{table*}

\begin{table*}[]
\small
\centering
\begin{tabular}{@{}lcccccccc@{}}
\toprule
          Similarity matching & \multicolumn{2}{c}{Accuracy} & \multicolumn{2}{c}{Precision} & \multicolumn{2}{c}{Recall} & \multicolumn{2}{c}{F1 Score} \\
\cmidrule(lr){2-3} \cmidrule(lr){4-5} \cmidrule(lr){6-7} \cmidrule(lr){8-9}
 & 4k & 12k & 4k &12k & 4k & 12k & 4k & 12k  \\
\midrule
Word2Vec + CNN    & 0.5016  & 0.5027 & 0.5086 & 0.5104 & 0.5016  & 0.5027  & 0.3646 & 0.3812 \\
\midrule
\textbf{BERT} & & & & & & & & \\
\midrule
VanillaBERT    & 0.8334  & 0.8444  & 0.8545& 0.8562  & 0.8334  & 0.8444 & 0.8304 & 0.8428  \\
\midrule
   Domain adapted & & & & & & & & \\
\midrule
$p_{nc}=0.50$ (MLM) & $0.8519^\dagger$ & $0.8639^\dagger$   & $\textbf{0.8641}^\dagger$  & $0.8746^\dagger$ & $0.8519^\dagger$ & $0.8639^\dagger$ & $0.8503^\dagger$ & $0.8627^\dagger$\\
$p_{nc}=0.75$ (LIM0.75, ours) & $\textbf{0.8574}^\dagger$ & $\textbf{0.8669}^\dagger$ & $0.8613^\dagger$ & $\textbf{0.8812}^\dagger$ & $\textbf{0.8574}^\dagger$ & $\textbf{0.8669}^\dagger$ & $\textbf{0.8568}^\dagger$ & $\textbf{0.8654}^\dagger$\\
$p_{nc}=1.00$ (LIM1, ours) & $0.8484^\dagger$  & $0.8599^\dagger$ & $0.8551^\dagger$ & $0.8724^\dagger$ & $0.8484^\dagger$ & $0.8599^\dagger$ & $0.8474^\dagger$ & $0.8584^\dagger$ \\
\midrule
\textbf{SciBERT} & & & & & & & & \\
\midrule
VanillaSciBERT & 0.8294& 0.8489& 0.8314& 0.8599& 0.8294& 0.8489& 0.8289& 0.8474\\
\midrule
   Domain adapted & & & & & & & & \\
\midrule
$p_{nc}=0.50$ (MLM)& $0.8524^\dagger$& $0.8684^\dagger$ & $\textbf{0.8733}^\dagger$ & $0.8808^\dagger$ & $0.8524^\dagger$& $0.8684^\dagger$& $0.8499^\dagger$ & $0.8671^\dagger$\\
$p_{nc}=0.75$ (LIM0.75, ours) & $\textbf{0.8614}^\dagger$& $\textbf{0.8689}^\dagger$& $0.8672^\dagger$& $\textbf{0.8827}^\dagger$& $\textbf{0.8614}^\dagger$& $\textbf{0.8689}^\dagger$ & $\textbf{0.8606}^\dagger$& $\textbf{0.8674}^\dagger$\\
$p_{nc}=1.00$ (LIM1, ours)& $0.8519^\dagger$ & $0.8664^\dagger$& $0.8711^\dagger$& $0.8774^\dagger$& $0.8519^\dagger$ & $0.8662^\dagger$& $0.8496^\dagger$& $0.8655^\dagger$\\
\bottomrule
\end{tabular}
\caption{Accuracy, precision, recall and F1-score of similarity matching on the test set for BERT and SciBERT without and with domain-adaptive pre-training with MLM or LIM ($p_{nc}=0.75/1.00$), $^\dagger$ indicates statistically significant difference to Vanilla baseline, $\alpha=0.05$}
\label{table:evalsim}
\end{table*}

\subsubsection{IPC classification}

For the IPC classification we use a subset of up to $480k$ labelled patent claims of the USPTO13M dataset for training, similar to \citet{lee2019patentbert}, and an test set of $150k$ patent claims, containing in total $738$ different IPC \cite{wipo_ipc} tags on the subclass level. We restrict our classification input to the claims as the input size of the model is limited and, as \cite{lee2019patentbert} have demonstrated, the text of the claims is sufficient to predict the IPC tags.

The patents have $1.73$ IPC tags on average. As the document class label we use the single most frequent tag after truncating all tags to the subclass level.
We remove tabs, multiple whitespace and mathematical formulas before passing the text to the BERT model for fine-tuning. For the IPC classification fine-tuning we choose the same hyperparameters as in the domain adaptive pre-training except the learning rate. For the learning rate we perform a grid search and choose the same learning rate of $5*10^{-5}$ for all configurations (see Appendix \ref{appendix:learningrateipc}).
We fine-tune in total for $30k$ steps, which corresponds to one epoch for $480k$ labelled samples, and we evaluate the models every $10k$ steps.

In order to analyze the impact of the language representations for downstream tasks with smaller number of labelled training data, we finetune the models on a training dataset size of $160k$ and $320k$ and the whole $480k$ samples and analyze the performance compared to the baseline.
Detailed results for the different dataset sizes for the BERT model as well as for the SciBERT model can be found in Table~\ref{table:evalipc}. Here the baseline performance is compared to the MLM or LIM domain adapted BERT and SciBERT model.

\subsubsection{Similarity matching}

For the similarity matching task we retrieve pairs of patents which stand in an ``X'' citation relation, which we interpret as indicating similarity between the two documents. We denote these pairs of patent which cite each other as positive pair. In order to fine-tune the models on similarity matching, we also need negative pairs of patent which do not stand in a citation relation. As the citations of a given patent do not include all possible true citable patents, but rather only those which the patent examiners choose, we must construct negative citation pairs.
This construction is done as follows: To a given patent document, which comes from the positive citation pairs, we sample randomly a negative patent document from the positive patent pairs which stand in a citation relation with another patent document.
If the pair turns out to be the same document, we drop it and also positive citations pairs are dropped.
We choose a training dataset size up of $12k$ citation pairs with $50.1\%$ positive and $49.9\%$ negative pairs represented by their claims. The test dataset contains $16,500$ pairs, $49.9\%$ positive and $50.1\%$ negative ones.
We remove tabs, multiple whitespaces and mathematical formulas from the text before passing it to the BERT model.

The grid search for the BERT vanilla model indicates that $2*10^{-5}$ is the most suitable learning rate for fine-tuning on similarity matching and we choose this rate for all configurations (see Appendix \ref{appendix:citationpredictionhyp}).

In order to analyze the impact of the language representations for downstream tasks with smaller number of labelled training data, we finetune the different configurations for one epoch on a training dataset size of $2k$, $8k$ and $12k$ samples and analyze the performance compared to the baseline.

The results of the evaluation for the similarity matching are provided in Table \ref{table:evalsim} for the BERT-based and for the SciBERT-based models.

\section{Results}

In the following we examine the downstream task evaluation results regarding our research questions.

\subsection{RQ 1: BERT vs SciBERT}

Comparing the evaluation results of BERT and SciBERT for IPC classification and for similarity matching leads to the conclusion that the SciBERT based models achieve an overall higher performance. For the IPC classification the results in Table \ref{table:evalipc} show that the SciBERT model outperforms the equally domain-adapted BERT model (for MLM, LIM0.75 and LIM1) by $1-2\%$ downstream task performance. In Table \ref{table:evalsim} we see the results for similarity matching and comparing the performance of the BERT-based and SciBERT-based models shows the same picture. The domain-adapted SciBERT based model outperforms the corresponding BERT model by $1-2\%$ downstream task performance. Overall the SciBERT model domain adapted with LIM with $p_{nc}=0.75$ achieves the best performance for IPC classification and similarity matching.

\subsection{RQ 2: MLM vs LIM}

We compare the domain-adaptive pre-training methods of MLM and LIM, by evaluating the downstream task performance of BERT and SciBERT domain adapted either with LIM with $p_{nc}=0.75$ or $p_{nc}=1.00$ or MLM.\newline
If we compare the evaluation results for the IPC classification in Table \ref{table:evalipc} we find that for BERT and SciBERT the domain adapted representations with LIM lead to a higher downstream task performance, for BERT LIM0.75 demonstrates the best results, for SciBERT LIM1 shows the highest performance. For the task of similarity matching we find similar results: BERT and SciBERT achieve the best performance for similarity matching based on the domain adapted representations of LIM0.75.

\begin{figure}
    \centering
    \includegraphics[width=0.5\textwidth]{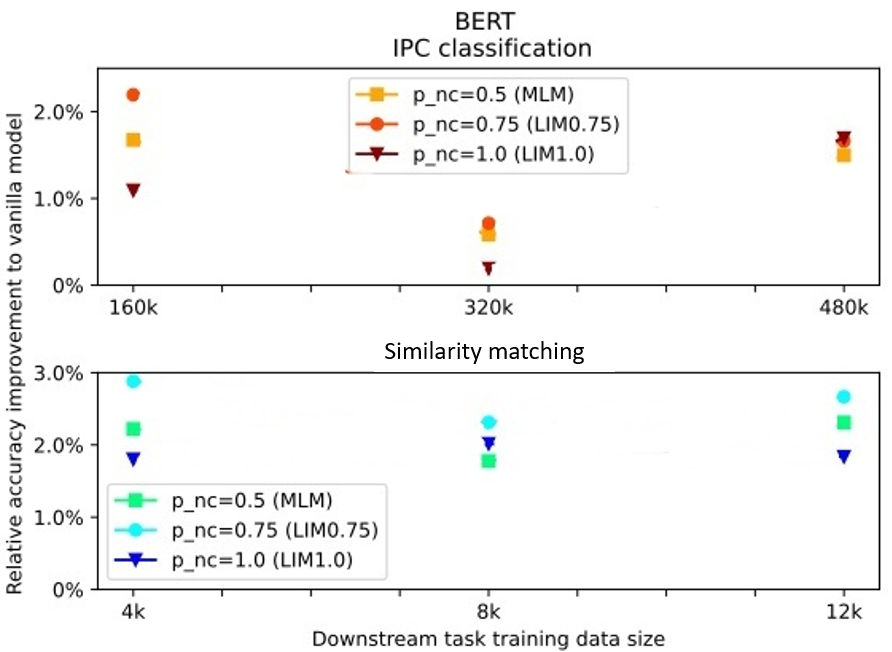}
        \caption{Relative accuracy improvement from vanilla BERT to MLM or LIM domain fine-tuned BERT models for different sizes of downstream task training data}
        \label{fig:relaccbert}
\end{figure}

\begin{figure}
    \centering
    \includegraphics[width=0.5\textwidth]{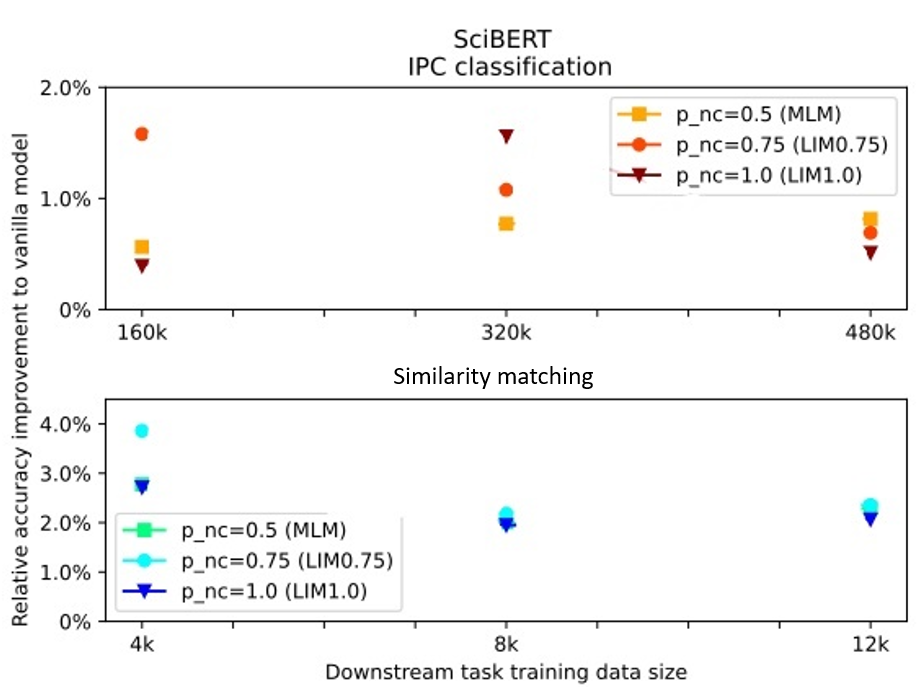}
        \caption{Relative accuracy improvement from vanilla SciBERT to MLM or LIM domain-finetuned SciBERT models for different sizes of downstream task training data}
        \label{fig:relaccscibert}
\end{figure}

In order to compare the performance gains to the baseline performance of the BERT models without domain adaptation, we can see the relative accuracy improvement compared to the baseline BERT model of the MLM and LIM domain fine-tuned models in Figures \ref{fig:relaccbert} and \ref{fig:relaccscibert} for the BERT-based models and the SciBERT-based models respectively.

For the BERT-based models we can see that the improvement of the LIM domain fine-tuned model with a noun chunk masking of $p_{nc}=0.75$ is consistently the highest for each size of downstream task training data for both downstream tasks, besides for the IPC classification trained on $480k$ samples, where the domain adapted model with $p_{nc}=1.00$ improves the performance compared to the domain adapted model with $p_{nc}=0.75$.
Similarly we observe that the LIM0.75 SciBERT model achieves the highest improvement for the similarity matching for all data set sizes, for the IPC classification the performance improvement for a smaller dataset size of $160k$ labelled samples is significant.

Especially in the setting of less training data for the downstream tasks, we can observe substantial performance improvements of the LIM0.75 domain fine-tuned models compared to the MLM fine-tuned model on both tasks and both models.
Therefore our experiments show that domain fine-tuning using LIM leads to improved representations, when comparing LIM to MLM on the two independent domain specific downstream tasks.\newline

These evaluation results demonstrate that the representations of patent language, which are domain adapted using the linguistically informed masking training method, achieve higher downstream task performance on two independent patent-related downstream tasks. Especially in the low data regime of the downstream task, the performance gains of the LIM representations are substantial.

\section{Analysis of results}

In this section we analyze our results regarding the better suitability of the SciBERT model to represent patent language than the BERT model as well as regarding the weighting factor $p_{nc}$ of the noun chunk masking in domain adaptive pre-training.

\subsection{Tokenization analysis}

The language models BERT and SciBERT are trained to encode a given text into representations. Every language model has a tokenization, and it has become common practice to train the tokenization of language models with a subword algorithm \cite{kudo2018sentencepiece}. The language model is then pre-trained in an unsupervised manner on domain language. Therefore the quality of the representations of a language model for a specific language domain depends on the similarity of the pre-training language and the target domain, as well as the suitability of the tokenization to the target domain.\newline
In this section we investigate one potential reason of the better suitability of the SciBERT model to encode patent language than of the BERT model: the tokenization. We analyze which tokenization is most suitable for patent language by comparing a subword tokenization trained on patent data to the tokenizations which BERT and SciBERT use for encoding.\newline
We measure the suitability of the tokenization by the split ratio, which is defined as length of the encoded sentence divided by the number of words in the sentence. A lower split ratio indicates better suitability for the domain because words are not split as often, thus domain-relevant words are retained in full more often rather than being split into less information rich parts.\newline
Following \citet{beltagy2019scibert} we train a vocabulary with the Sentencepiece algorithm \cite{kudo2018sentencepiece} on $5$ million sentences of the patent dataset.
As the split ratio of the training sentences is similar to the split ratio of unseen patent sentences, we conclude that the tokenization is sufficiently well trained on these $5$ million training sentences to be a general encoding of patent language.
Now we want to compare the different vocabularies for encoding patent language. 
The example in Table~\ref{table:vocab} shows how a suitable tokenization leads to less splitting up into subwords as it includes special words such as ``femto'' in full.
\begin{table}
\begin{tabular}{@{}ll@{}}
\toprule
Tokenization & Encoding \\
\midrule
BERT & [\textcolor{orange}{ 'f', '\#\#em', '\#\#to',} 'access', 'point'] \\
SciBERT & [\textcolor{orange}{'fem', '\#\#to',} 'access', 'point'] \\
Patent & [\textcolor{orange}{'femto',} 'access', 'point']\\
\bottomrule
\end{tabular}
\caption{Encoding of the fragment ``femto access point'' using three different trained vocabularies}
\label{table:vocab}
\end{table}

\begin{figure}
    \centering
    \includegraphics[width=0.50\textwidth]{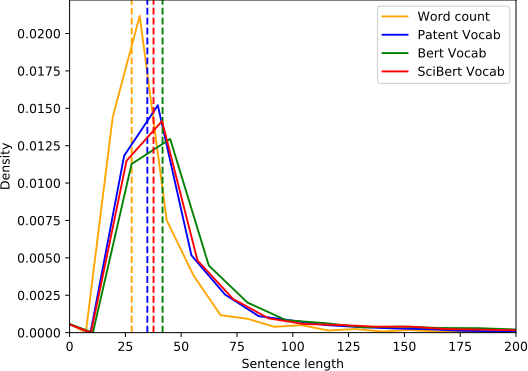}
    \caption{Distribution of the sentence length for different tokenizations}
    \label{fig:vocabcomparison}
\end{figure}
In order to compare the performance of the different vocabularies for encoding patent language we determine the length of the encodings for $1.6$ million sentences from patents with the different vocabularies.\newline
In Figure \ref{fig:vocabcomparison} is the distribution of the sentence length in total number of words and the distribution of the different encoding lengths of the sentences with the average as dashed vertical line.

We can see that the encoding length with the SciBERT tokenization is shorter than the encoding with the BERT vocabulary. 
We observe an average split ratio for the patent tokenization of $1.16$, for the SciBERT tokenization of $1.21$ and for the BERT tokenization of $1.29$. Because of the shorter encoding length and the smaller split ratio of the SciBERT tokenization relative to BERT, we can conclude that the SciBERT tokenization fits better to encode the patent language.
Because the representations are learned for the tokens in the vocabulary, it follows that the better the tokenization fits to the language domain the more specific information can be captured by the learned representations.
As the tokenization of the SciBERT language model fits better to represent patent language than the BERT vocabulary, we suggest this as a reason that the SciBERT language model shows better results for representing patent language than BERT.

\subsection{Masking probability analysis}

\begin{figure}
    \centering
    \includegraphics[width=0.5\textwidth]{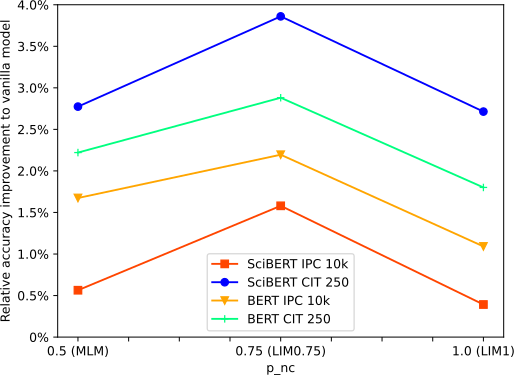}
    \caption{Effect of balancing the noun chunk masking with the parameter $p_{nc}$: representations which are domain fine-tuned using LIM0.75 show promising accuracy improvements for both downstream tasks and for both BERT-based representations}
        \label{fig:balancenounchunks}
\end{figure}

We also want to investigate the effect of balancing the domain adaptive pre-training with the noun chunk masking parameter $p_{nc}$ which gives the ability to control the learning from different linguistic information from the target domain. For that we analyze the accuracy improvements of different noun chunk masking values $p_{nc}$ for domain adaptive pre-training for BERT and SciBERT on both downstream tasks.

We show the accuracy improvements compared to the baseline models without domain adaptative pre-training in Figure \ref{fig:balancenounchunks}.
Here one can observe the clear trend that shifting the noun chunk masking towards the noun chunks with a weighting of $75\%$ masked noun chunks and $25\%$ masked non-noun chunks lead to higher accuracy improvements for both downstream tasks. However focusing the domain fine-tuning only on the noun chunks, in other words LIM1, shows suboptimal results, which leads to the conclusion that balancing the masking of noun chunks and non-noun chunks between the values of $0.5$ and $1.0$ is beneficial for domain adaptive pre-training for patent language.


\section{Conclusion}


Overall we conclude that domain adaptive pre-training for learning representations of patent language is beneficial for pre-trained masked language models like BERT and SciBERT. We find that the SciBERT-based representations outperform the BERT-based representations of patent language for the two independent downstream tasks of IPC classification and similarity matching. Therefore we conclude that SciBERT is more suitable to represent patent language than the BERT model and see one reason for that in the suitability of the tokenization of SciBERT to patent language. Furthermore we have proposed the empirically motivated domain adaptive pre-training method of linguistically informed masking for BERT-like language models. We demonstrate improvements on both patent-related downstream tasks for representations of patent language that have been domain-adapted using the LIM method. Furthermore we analyze the impact of the weighting factor for shifting the masking towards the noun chunks.
We conclude that domain adaptive pre-training with linguistically informed masking improves the representations of the patent domain for BERT and SciBERT and that balancing the weighting to learn from different linguistic information is beneficial for representation learning.


\bibliographystyle{ACM-Reference-Format}
\bibliography{sample-base}

\appendix

\section{Appendices}
\label{sec:appendix}
\subsection{Google BigQuery query for patent dataset USPTO13M}
\label{appendix:a}

\begin{lstlisting}[
           language=SQL,
           showspaces=false,
           basicstyle=\ttfamily,
           %numbers=left,
           numberstyle=\tiny,
           commentstyle=\color{gray}
        ]
SELECT
publication_number,
publication_date,
filing_date,
priority_date,
title.text AS title,
title.truncated as title_tr,
abstract.text AS abstract,
abstract.truncated as abstract_tr,
claim.text as claim,
claim.truncated as claim_tr,
descr.text as descr,
descr.truncated as descr_tr,
ARRAY_TO_STRING(ARRAY(SELECT
code FROM UNNEST(p.ipc)), ";") 
AS ipc,
ARRAY_TO_STRING(ARRAY(SELECT
code FROM UNNEST(p.cpc)), ";") 
AS cpc
FROM
`patents-public-data.patents.
publications` p,
UNNEST(p.title_localized) as title,
UNNEST(p.abstract_localized) 
as abstract,
UNNEST(p.claims_localized) as claim,
UNNEST(p.description_localized)
as descr
WHERE
filing_date >= 20000101
AND
claim.language = 'en'
AND
descr.language = 'en'
AND
title.language = 'en'
AND
abstract.language = 'en'
AND
/* Granted patents only */
application_kind = 'A'
\end{lstlisting}

Google BigQuery database accessed on the 04.11.2019\newline

\subsection{Hyperparameter for domain adaptive pre-training on patent domain}
\label{appendix:domainfinetunelearningrate}

The masking and next sentence accuracies after domain adaptive pre-training each configuration for $2500$ steps for the different learning rate candidates $1*10^{-4}, 5*10^{-5}, 2*10^{-5}, 1*10^{-5}$ as well as the learning rate choice for each configuration can be seen in Table \ref{table:furtherpretrainconfigandhyper}.
Domain adaptive pre-training was performed on $1$ GPU and took $2.5$ days for each configuration.

\begin{table}[h]
\centering
    \begin{tabular}{l l}
    \toprule
    maximum sequence length & 128 \\
    masking probability & 0.15 \\
    training steps & 100,000  \\ 
    warm-up steps & 10,000 \\ 
    $\beta_1$ & 0.9  \\ 
    $\beta_2$ & 0.999 \\ 
    dropout probability & 0.1 \\
    batch size & 32 \\
    \bottomrule
    \end{tabular}
\caption{Hyperparameter for domain adaptive pre-training}
\label{table:furtherpretrainconfigandhyper}
\end{table}

\subsection{Independence of IPC classification and similarity matching}
\label{appendix:independeceipccit}

In order to test whether there is a trivial relationship between the IPC tags and the similarity relations of a patent, we train a linear SVM classifier on predicting the similarity matching of a pair of patents from their IPC tag representation. On an equally balanced binary dataset we reach a classification performance which is little better than random (accuracy: 0.59, F1 score: 0.46), and therefore conclude that the two tasks are independent.

\newpage 
\subsection{Learning rate evaluation for fine-tuning on IPC classification}
\label{appendix:learningrateipc}

The evaluation of the grid search for the best suitable learning rate for IPC classification fine-tuning can be found in Table \ref{table:ipclearningrateexp}. The accuracy values for the different configurations are shown after fine-tuning for $2500$ steps with the learning rate candidates $5*10^{-5}, 2*10^{-5}, 1*10^{-5}$.
The IPC classification was performed on 1 GPU and took around 1.5 days for each configuration.

\begin{table}[h]
\small
\centering
    \begin{tabular}{lccc}
    \toprule
    & \multicolumn{3}{c}{IPC classification} \\
    \cmidrule(lr){2-4} 
     & \textbf{$5*10^{-5}$} & $3*10^{-5}$ & $2*10^{-5}$  \\ 
    \midrule
    BERT Vanilla & \textbf{0.4321} & 0.4301 & 0.4135  \\ 
    BERT MLM & \textbf{0.4737}  & 0.4598 & 0.4333 \\ 
    BERT LIM0.75 & \textbf{0.4776}  & 0.4594 & 0.4372\\ 
    BERT LIM1 & \textbf{0.4830}  & 0.4592 & 0.4212 \\ 
    SciBERT Vanilla & \textbf{0.4906}  & 0.4773 & 0.4501 \\ 
    SciBERT MLM & \textbf{0.5031}  & 0.48705 & 0.4652 \\ 
    SciBERT LIM0.75 & \textbf{0.5142}  & 0.4870 & 0.4665  \\ 
    SciBERT LIM1 & \textbf{0.5020}  & 0.4863 & 0.4647 \\ 
    \bottomrule
    \end{tabular}
\caption{Accuracy values for different learning rates after IPC classification fine-tuning each model for $2500$ steps}
\label{table:ipclearningrateexp}
\end{table}

\subsection{Hyperparameter for fine-tuning on similarity matching}
\label{appendix:citationpredictionhyp}

The hyperparameter for fine-tuning on similarity matching can be found in Table \ref{table:citpredfinetuninghyp}. Fine-tuning the BERT vanilla model configuration for $1000$ steps on the different learning rates of $5*10^{-5}$, $3*10^{-5}$, $2*10^{-5}$ indicates that $2*10^{-5}$ is the most suitable learning rate for fine-tuning on similarity matching and we choose this rate for all configurations.
The fine-tuning was performed on $1$ GPU with $61$ RAM and took around $30$ hours for each configuration.

\begin{table}[h]
\centering
    \begin{tabular}{l l}
    \toprule
    maximum sequence length & 256 \\
    masking probability & 0.15 \\
    training steps & 1000  \\ 
    warm-up steps & 100 \\ 
    $\beta_1$ & 0.9  \\ 
    $\beta_2$ & 0.999 \\ 
    dropout probability & 0.1 \\
    batch size & 16 \\
    learning rate & $2*10^{-5}$ \\
    \bottomrule
    \end{tabular}
\caption{Hyperparameter for similarity matching fine-tuning}
\label{table:citpredfinetuninghyp}
\end{table}

\end{document}